\documentclass[11pt,a4paper]{article}
\usepackage[hyperref]{emnlp-ijcnlp-2019}
\usepackage{times}
\usepackage{latexsym}
\usepackage{latexsym}
\usepackage{graphicx}
\usepackage{amsmath}
\usepackage{mathtools}
\usepackage{breqn}
\usepackage{amsmath}
\usepackage{url}
\usepackage{enumerate}
\usepackage{amsmath}
\usepackage{amssymb}
\usepackage{comment}
\usepackage{xspace}

\makeatletter
\DeclareRobustCommand\onedot{\futurelet\@let@token\@onedot}
\def\@onedot{\ifx\@let@token.\else.\null\fi\xspace}

\def\eg{\emph{e.g}\onedot} 
\def\eg{\emph{e.g}\onedot} 
\def\ie{\emph{i.e}\onedot}

\aclfinalcopy 

\title{Semantic Relatedness Based Re-ranker for Text Spotting}

\author{Ahmed Sabir$^{1}$ \hspace{0.7cm}
Francesc  Moreno-Noguer$^{2}$ \hspace{0.7cm}
 Llu\'{\i}s Padr\'o$^{1}$ \\
$^{1}$ Universitat Polit\`ecnica de Catalunya, TALP Research Center, Barcelona, Spain  \\ 
$^{2}$ Institut de Rob\`otica i Inform\`atica Industrial, CSIC-UPC, Barcelona, Spain\\
{\tt asabir@cs.upc.edu, fmoreno@iri.upc.edu, padro@cs.upc.edu}
}
\date{}

\begin{document}
\maketitle
\begin{abstract}
Applications such as textual entailment, plagiarism detection or document clustering rely on the notion of \textit{semantic similarity}, and are usually approached with dimension reduction techniques like LDA or with embedding-based neural approaches. We present a scenario where semantic similarity is not enough, and we devise a neural approach to learn \textit{semantic relatedness}. The scenario is \textit{text spotting in the wild},  where a text in an image (\eg street sign, advertisement or bus destination) must be identified and recognized. Our goal is to improve the performance of vision systems by leveraging semantic information. Our rationale is that the text to be spotted is often related to the image context in which it appears (word pairs such as \textit{Delta}--\textit{airplane}, or \textit{quarters}--\textit{parking} are not \textit{similar}, but are clearly \textit{related}). We show how learning a word-to-word or word-to-sentence relatedness score can improve the performance of text spotting systems up to 2.9 points, outperforming other measures in a benchmark dataset. 

\end{abstract}

\section{Introduction}
Deep learning has been successful in tasks related to deciding whether two short pieces of text refer to the same topic, \eg semantic textual similarity \citep{Daniel:18}, textual entailment  \citep{Ankur:16} or answer ranking for Q\&A \citep{Aliaksei:15}. 

However, other tasks require a broader perspective to decide whether two text fragments are \textit{related} more than whether they are \textit{similar}. In this work, we describe one of such tasks, and we retrain some of the existing sentence similarity approaches to learn this semantic relatedness. We also present a new Deep Neural Network (DNN)  that outperforms existing approaches when applied to this particular scenario.

The task we tackle  is \textit{Text Spotting},  which is the problem of recognizing text that appears in unrestricted images (a.k.a. \textit{text in the wild}) such as traffic signs, commercial ads, or shop names. Current state-of-the-art results on this task are far from those of OCR systems with simple backgrounds. 

Existing approaches to Text Spotting usually divide the problem in two fundamental tasks: 1) \textit{text detection}, consisting of selecting the image regions likely to contain texts, and 2) \textit{text recognition}, that converts the images within these bounding boxes into a readable string. In this work, we focus on the recognition stage, aiming to prove that semantic relatedness between the image context and the recognized text can be useful to boost the system performance. We use existing pre-trained architectures for Text Recognition, and add a shallow deep-network that performs a  post-processing operation to re-rank the proposed candidate texts. In particular, we re-rank the candidates using their semantic relatedness score with other visual information extracted from the image (\eg objects, scenario, image caption). Extensive evaluation shows that our approach consistently improves other semantic similarity methods.

\begin{figure*}[t!]
 \centering 
\includegraphics[width=5.4in]{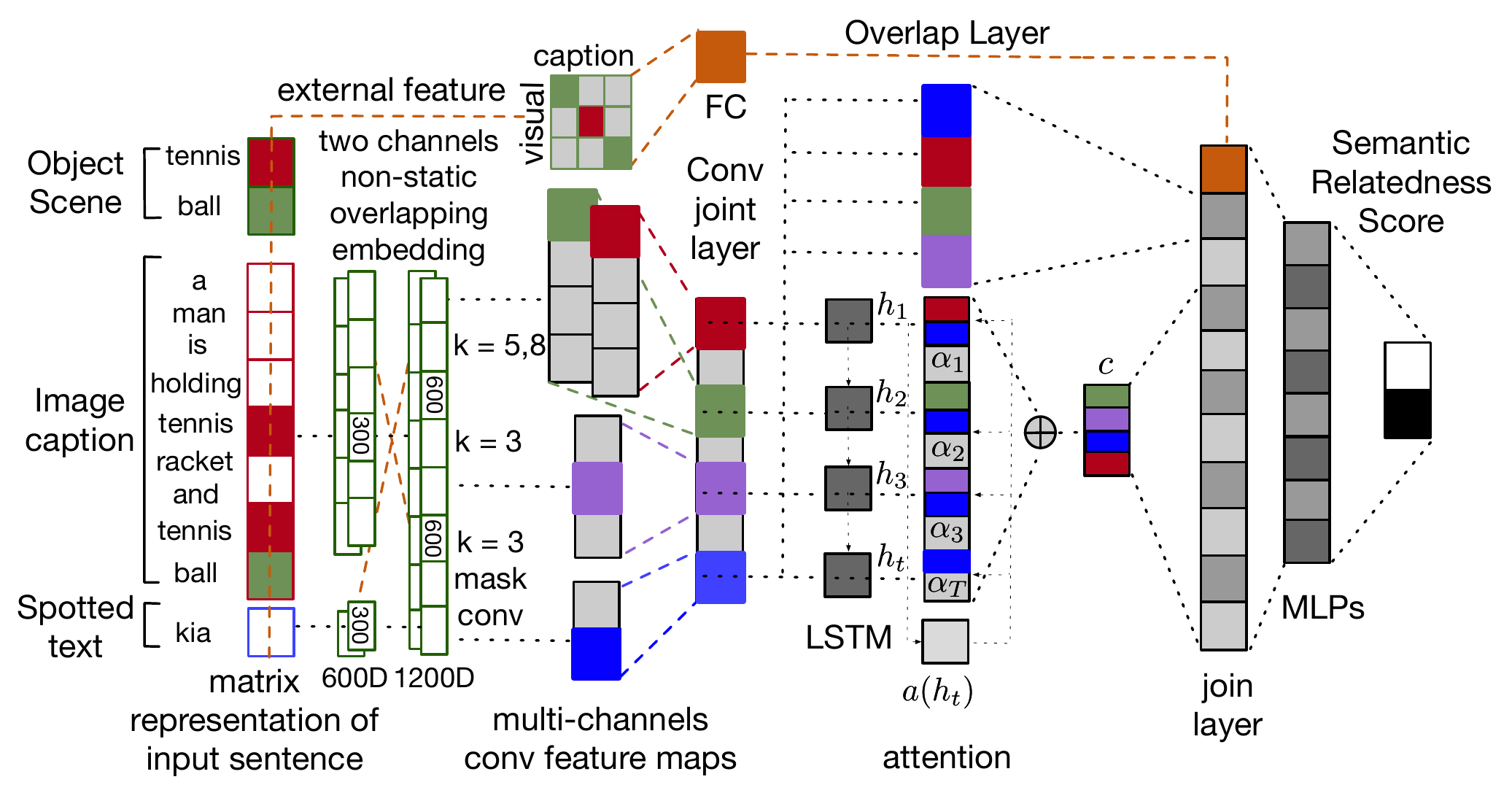}
\caption{Overview of the system pipeline, an end-to-end post-process scores the semantic relatedness between a candidate word and the context in the image (objects, scenarios, natural language descriptions, ...)}
\label{fig:fcnn}
\end{figure*}

\section{Text Hypothesis Extraction }
We use two pre-trained Text Spotting baselines to extract $k$ text hypotheses. The first baseline is a CNN \citep{Max:16} with fixed lexicon based recognition, able to recognize words in a predefined 90K-word dictionary. Second, we use an LSTM architecture with visual attention model \citep{Suman:17} that generates the final output words as probable character sequences, without relying on any lexicon. Both models are trained on a synthetic dataset \citep{Max:14b}. The output of both models is a Softmax score for each of the $k$ candidate words.

\section{Learning Semantic Relatedness for Text Spotting} 
\label{sec:re-ranking-word}
To learn the semantic relatedness between the visual context information and the candidate word we introduce a multi-channel convolutional LSTM with  an attention mechanism.
The network is fed with the candidate word plus several words describing the image visual context (object and places labels, and descriptive captions)\footnote{All this visual context information is automatically generated using off-the-shelf existing modules (see section \ref{sec:data}).}, and is trained to produce a relatedness score between the candidate word and the context.


Our architecture is inspired by \citep{Aliaksei:15}, that proposed   CNN-based re-rankers for Q\&A. Our network consists of two subnetworks, each with 4-channels with kernel sizes $k=(3,3,5,8)$, and overlap layer, as shown in Figure \ref{fig:fcnn}. We next describe the main components:

\noindent{\bf{Multi-Channel Convolution}}: The first subnetwork consists of only  convolution kernels, and aims to extract n-gram or keyword features from the caption sequence. 
The convolution is applied over a sequence to extract n-gram features from different positions. Let $x \in \mathbb{R}^{s \times d}$  be the sentence matrix, where  $s$ is  the sentence length, and $d$ the  dimension of the i-th word in the sentence. Let also denote by   $c \in \mathbb{R}^{k\times {d}}$ the kernel for the convolution operation.  For each $i$-th position   in the sentence,   $w_{i}$ is the concatenation of  $k$ consecutive words, i.e.,  $w_{i} = [x_{i} \oplus x_{i+1} \oplus \dots \oplus x_{i+k-1}]$.
Our architecture uses multiple such kernels to generate feature maps $m$. The feature map for each window vector $w_{i}$ can be written as:
\begin{equation}
m_{i} = f(w_{i} \circ c + b) 
\end{equation}
where $\circ$ is element-wise multiplication, $f$ is nonlinear function, in our case we apply Relu function \citep{Vinod:10}, and $b$ is a bias. For $j$ kernels, the generated $j$ feature maps can be arranged as feature representation for each window $W_{i}$ as: $W = [m_{1} \oplus m_{2} \oplus \ldots \oplus m_{j}]$. Each row $W_{i}$ of $W \in \mathbb{R}^{(s-k+1)\times {j}}$ is the new generated feature from the $j$-th kernel for the window vector at position $i$. The new generated feature (window representations)  are then fed into the joint layer and LSTM  as shown in Figure \ref{fig:fcnn}.

\noindent{\bf{Multi-Channel Convolution-LSTM}}: Following C-LSTM \citep{Chunting:15} we forward the output of the CNN layers  into an LSTM, which captures the long term dependencies over the features. We further introduce an attention mechanism to capture the most important features from that sequence. The advantage of this attention is that the model learns the sequence without relying on the temporal order. We describe in more detail the attention mechanism below.

Also, following \citep{Chunting:15}, we do not use a pooling operation after the convolution feature map. Pooling layer is usually applied after the convolution layer to extract the most important features in the sequence. However, the output of our Convolutional-LSTM model is fed into an LSTM \citep{Sepp:97} to learn the extracted sequence, and pooling layer would break that sequence via downsampling to a selected feature. In short, LSTM is specialized in learning sequence data, and pooling operation would break such a sequence order. On the other hand, for the Multi-Channel Convolution model we also lean the extracted word sequence n-gram directly and without feature selection, pooling operation.


\noindent{\bf{Attention Mechanism}}: Attention-based models have shown promising results on various NLP tasks \citep{Dzmitry:14}. Such mechanism learns to focus  on a specific part of the input (\eg a relevant word in a sentence).
We apply an attention mechanism \citep{Colin:15} via an LSTM that captures the temporal dependencies in the sequence. This attention uses a Feed Forward Neural Network (FFN) attention function:
\begin{equation}
e_{t}= \tanh \left(h_{t} W_{a}\right) v_{a}^{\mathrm{T}} 
\end{equation}
\noindent where $W_{a}$ is the attention of the hidden weight matrix and $v_{a}$ is the output vector.  As shown in Fig. \ref{fig:fcnn} the vector $c$ is computed as a weighted average of $h_{t}$, given by $\alpha$ (defined below). The attention mechanism is used to produce a single vector $c$ for the complete sequence as follows: 
\begin{equation*}
e_{t}=a\left(h_{t}\right), 
\alpha_{t}=\frac{\exp \left(e_{t}\right)}{\sum_{k=1}^{T} \exp \left(e_{k}\right)}, c=\sum_{t=1}^{T} \alpha_{t} h_{t} 
\end{equation*}

where $T$ is the total number of steps and $\alpha_{t}$ is the computed weight of each time step $t$ for each state $h_{t}$, $a$ is a learnable function that depends only on $h_{t}$. Since this attention  computes the average over time, it discards the temporal order, which is ideal for learning semantic relations between words. By doing this, the attention gives higher weights to more important words in the sentence without relying on sequence order.

\noindent{\bf{Overlap Layer}:} The overlap layer is just a frequency count dictionary to compute   overlap information of the inputs. The idea is to give more weight to the most frequent visual element, specially when it is observed by more than one visual classifier. The dictionary output is a fully connected layer.  


Finally, we merge all subnetworks into a joint layer that is fed to a loss function which calculates the semantic relatedness between both inputs. We call the combined model Fusion Dual Convolution-LSTM-Attention (FDCLSTM$_{AT}$). 

 Since we have only one candidate word at a time, we apply a convolution with \textit{masking} in the candidate word side (first channel). In this case, simply zero-padding the sequence has a negative impact on the learning stability of the network. We concatenate the CNN outputs with the additional feature into MLP layers, and finally a sigmoid layer performing binary classification. We trained the model with a binary cross-entropy loss ($l$) where the target value (in $[0,1]$) is the semantic relatedness between the word and the visual. Instead of restricting ourselves to a simple similarity function, we let the network learn the margin between the two classes --\ie the degree of similarity. For this, we increase the depth of network after the MLPs merge layer with more fully connected layers. The network is trained using Nesterov-accelerated Adam (Nadam) \citep{Timothy:16} as it yields  better results  (specially in cases such as word vectors/neural language modelling) than other optimizers using only classical momentum (ADAM). We apply batch normalization (BN) \citep{Sergey:15} after each convolution, and between each MLPs layer. We omitted the BN after the convolution for the model without attention (FDCLSTM), as BN deteriorated the performance.  Additionally, we consider 70\% dropout \citep{Nitish:14} between each MLPs for regularization purposes.

\begin{table*}[t!]
\small
\caption{Best results after re-ranking using different re-ranker, and different values for $k$-best hypotheses extracted from the baseline output (\%). In addition, to evaluate our re-ranker with MRR we fixed  $k$ CNN$_{k=8}$ LSTM$_{k=4}$} 
\centering
\begin{tabular}{|l|c c c c c |c c c c |} 

\hline 
 \textbf{Model}  & \multicolumn{5}{c|}{\textbf{CNN}}   & \multicolumn{4}{c|}{\textbf{LSTM}}  \\ 
  
  & \textit{full} & \textit{dict} & \textit{list} &  \textit{k} &  MRR &   \textit{full} &  \textit{list} & k & MRR \\ 

\hline
\hline
Baseline (BL)  & \multicolumn{5}{c|}{\textbf{full: 19.7 dict: 56.0}}   & \multicolumn{4}{c|}{\textbf{full: 17.9}}  \\
\hline 

BL+ Glove      \citep{Jeffrey:14}    &   22.0    &   62.5     & 75.8   &  7   & 44.5    & 19.1        & 75.3    & 4  & 78.8               \\
BL+C-LSTM \citep {Chunting:15}          & 21.4       &   61.0 &71.3   &   8  &45.6    &  18.9   &  74.7    & 4 &  80.7   \\ 
BL+CNN-RNN \citep{Xingyou:16}           & 21.7      &   61.8 & 73.3     &  8  & 44.5    &  19.5   &  77.1  & 4  & 80.9  \\ 
BL+MVCNN \citep{Wenpeng:16}                & 21.3 &  60.6  &71.9  & 8     & 44.2 & 19.2 & 75.8 & 4   & 78.8 \\
BL+Attentive LSTM \citep{Ming:16}               &   21.9     &  62.4  &  74.0 & 8  & 45.7   &   19.1 & 71.4     & 5 & 80.2 \\
BL+fasttext \citep{Armand:17}        &   21.9    &   62.2     &  75.4     &  7   & 44.6 & 19.4           &  76.1    & 4       & 80.3       \\ 
BL+InferSent \citep{Alexis:17}        &   22.0      &   62.5 & 75.8 &    7 & 44.5   &  19.4  & 76.7    &   4 &  79.7                     \\
BL+USE-T \citep{Daniel:18}         &   22.0       &    62.5     &  \textbf{78.3}  &   6  & 44.7   & 19.2    &  75.8  & 4    & 79.5       \\ 
BL+TWE     \citep{Ahmed:18}         &   22.2    &   63.0     & 76.3   &  7   & 44.7 & 19.5        & 76.7 & 4     &80.2                            \\
BL+FDCLSTM            (ours)                  &  22.3     &   63.3 &  75.1   & 8    & 45.0 &  \textbf{20.2}    & 67.9  & 9 & 79.8 \\    
BL+FDCLSTM$_{AT}$  (ours) &  \textbf{22.4}    &    63.7  & 75.5    & 8 & \textbf{45.9}  & 20.1  & 67.6  &  9  & \textbf{81.8}              \\ 
\hline 
\hline
BL+FDCLSTM$_{lexicon}$ (ours)   &  22.6    &   64.3         &  76.3  &  8  & 45.1 & 19.4  & 76.4  &  4 & 78.8  \\
BL+FDCLSTM$_{AT}$$_{+}$$_{lexicon}$  (ours) &  22.6    &  64.3      &  76.3   & 8 & 45.1 & 19.7  & \textbf{77.8}  &  4  & 80.4     \\ 

\hline 

\end{tabular}
\footnotetext{}
\label{table_results}
\end{table*}

\section{Dataset and Visual Context Extraction }
\label{sec:data}
\begin{figure*}[t!]
\centering 
\includegraphics[width=\textwidth]{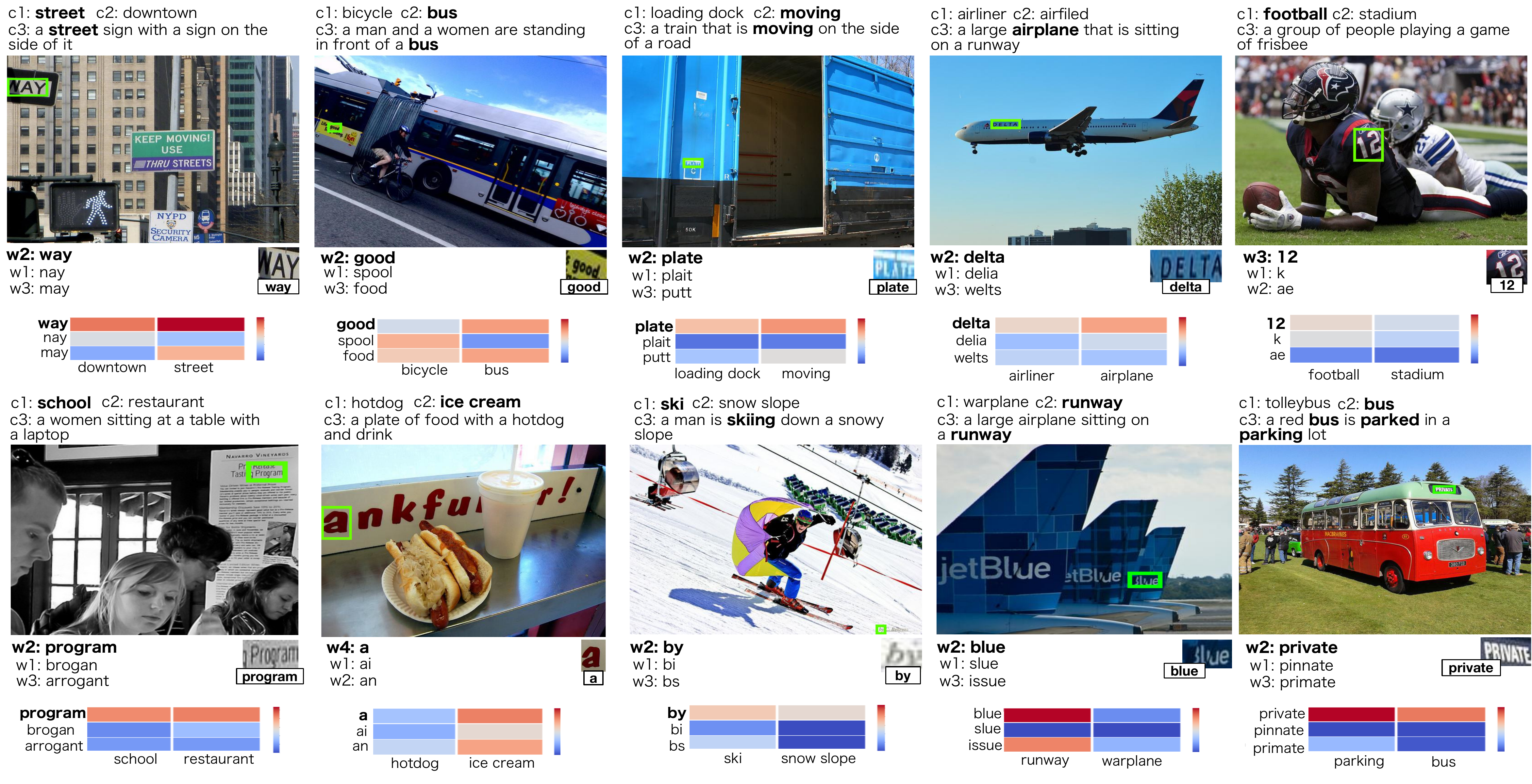} 
\vspace{-8mm}
\caption{Examples of candidate re-ranking using object (c1), place (c2), and caption (c3) information. The top three examples are re-ranked  based on the semantic relatedness score. The \textit{delta}-\textit{airliner} which frequently co-occur in training data is captured by overlap layers. The \textit{12}-\textit{football} show that the relation between sport and numbers. Also, \textit{program}-\textit{school} have a much more distance relation but our model is able to re-rank the most related words. The \textit{blue-runway} and \textit{bus-private} show that the overlap layer can be effective when the visual context appears in more than one visual context classifier. Finally, \textit{hotdog}-\textit{a} and \textit{by}-\textit{ski} have no semantic correlation but are solved by network thanks to the frequency count dictionary.}
 \label{fig:caption}
 \end{figure*}

We evaluate the performance of the proposed approach on the COCO-text \citep{Andreas:16}.
This dataset is based on Microsoft COCO \citep{Tsung-Yi:14} (Common Objects in Context), which consists of 63,686 images, and 173,589 text instances (annotations of the images). This dataset does not include any visual context information, thus we used out-of-the-box object  \citep {Kaiming:16} and place \citep{Bolei:14} classifiers and tuned a caption generator \citep{Oriol:15} on the same dataset to extract contextual information from each image, as seen in Figure~\ref{fig:caption}.

\section{Related Work and Contribution}
Understanding the visual environment around the text is very important for scene understanding. This has been recently explored by a relatively reduced number of works. \citet{Anna:16} shows that the visual context could be beneficial for text detection. This work uses a 14 classes pixel classifier to extract \textit{context features} from the image, such as \textit{tree, river, wall}, to then assist scene text detection. \citet{Chulmoo:17} employs topic modeling to learn the correlation between visual context and the spotted text in social media. The metadata associated with each image (e.g tags, comments and titles) is then used as context to enhance recognition accuracy. \citet{Sezer:17} takes advantage of text and visual context for logo retrieval problem. Most recently, \citet{Shitala:18} use object information (limited to 42 predefined object classes) surrounding the spotted text to guide text detection. They propose two sub-networks to learn the relation between text and object class (e.g. relations such as \textit{car}--\textit{plate} or \textit{sign board}--\textit{digit}).

Unlike these methods, our approach uses direct visual context  from  the  image  where  the  text  appears,  and does not rely on any extra resource such as human labeled meta-data \citep{Chulmoo:17} nor limits the context object classes  \citep{Shitala:18}. In addition, our approach is easy to train and can be used as a drop-in complement for any text-spotting algorithm that outputs a ranking of word hypotheses. 

\section{Experiments and Results}
\label{sec:experiments}
In the following we use different \textit{similarity} or \textit{relatedness} scorers to reorder the $k$-best hypothesis produced by an off-the-shelf state-of-the-art text spotting system. We experimented extracting  $k$-best hypotheses for $k=1\ldots10$.

We use two pre-trained deep models: a CNN \citep{Max:16} and an LSTM \citep{Suman:17} as baselines (BL) to extract the initial list of word hypotheses.

The CNN baseline uses a closed lexicon; therefore, it cannot recognize any word outside its 90K-word dictionary.  Table~\ref{table_results} presents four different accuracy metrics for this case: 1) \textbf{full} columns correspond to the accuracy on the whole dataset. 2) \textbf{dict} columns correspond to the accuracy over the cases where the target word is among the 90K-words of the CNN dictionary (which correspond to 43.3\% of the whole dataset. 3) \textbf{list} columns report the accuracy over the cases where the right word was among the $k$-best produced by the baseline. 4) \textbf{MRR} Mean Reciprocal Rank (MRR), which is computed as follows: $\mathrm{MRR}=\frac{1}{|Q|} \sum_{k=1}^{|Q|} \frac{1}{\operatorname{rank}_{k}}$, where rank $k$ is the position of the correct answer in the hypotheses list proposed by the baseline.

\noindent{\bf{Comparing with sentence level model}:} We compare the results of our encoder with several state-of-the-art sentence encoders, tuned or trained on the same dataset. We use cosine to compute the similarity between the caption and the candidate word. Word-to-sentence representations are computed with:  Universal Sentence Encoder with the Transformer USE-T \citep{Daniel:18}, and Infersent \citep{Alexis:17} with glove \citep{Jeffrey:14}. The rest of the systems in Table~\ref{table_results} are trained in the same conditions that our model with glove initialization with dual-channel overlapping 
non-static pre-trained embedding on the same dataset. Our model FDCLSTM without attention achieves a better result in the case of the second baseline LSTM that full of false-positives and short words. The advantage of the attention mechanism is the ability to integrate information over time, and it allows the model to refer to specific points in the sequence when computing its output. However, in this case, the attention attends the wrong context, as there are many words have no correlation or do not correspond to actual words. On the other hand, USE-T seems to require a shorter hypothesis list to get top performance when the right word is in the hypothesis list.



\noindent{\bf{Comparing with word level model}:} We also compare our result with current state-of-the-art word embeddings trained on a large general text using  \textit{glove} and \textit{fasttext}. The word model used only object and place information, and ignored the caption. Our  proposed models achieve better performance than  our TWE previous model \citep{Ahmed:18}, that trained a word embedding \citep{Tomas:13} from scratch on the same task.

\noindent{\bf{Similarity to probabilities}:} After computing the cosine similarity we need to convert that score to probabilities. As we proposed in previous work \citep{Ahmed:18} we obtain the final probability combining \citep{Sergey:03} the similarity score, the probability of the detected context (provided by the object/place classifier), and the probability of the candidate word (estimated from a 5M token corpus) \citep{Pierre:16}. 

\noindent{\bf{{Effect of Unigram Probabilities:}}} ~\citet{Suman:17} showed the utility of a language model (LM) when the data is too small for a DNN, obtaining significant improvements. Thus, we introduce a basic model of unigram probabilities with Out-of-vocabulary (OOV) words smoothing. The model is applied at the end, to re-rank out false positive short words, and has the main goal of re-ranking out less probable word overranked by the deep model.  As seen in Table \ref{table_results}, the introduction of this unigram lexicon produces the best results.

\noindent{\bf{Human performance}}: To estimate an upper bound for the results, we picked 33 random pictures from the test dataset and had 16 human subjects try to select the right word among the top $k=5$ candidates produced by the baseline text spotting system. Our proposed model performance on the same images was 57\%. Average human performance was 63\% (highest 87\%, lowest 39\%).

\section{Conclusion}
 In this work, we propose a simple deep learning architecture  to learn semantic relatedness between word-to-word and word-to-sentence pairs, and show how it outperforms other semantic similarity scorers when used to re-rank candidate answers in the Text Spotting problem.

In the future, we plan using the same approach to tackle similar  problems, including  lexical selection in Machine Translation, or word sense disambiguation \citep{Chiraag:18}. We believe our approach  could  also be useful in multimodal machine translation, where an image caption must be translated using not only the text but also the image content~\cite{barrault2018findings}. Tasks that lie at the intersection of computer vision and NLP, such as the challenges posed in the new BreakingNews dataset (popularity prediction, automatic text illustration) could also benefit from our results~\cite{Ramisa_pami2018}. 

\section*{Acknowledgments}
We would like to thank Jos\'e Fonollosa,  Marta Ruiz Costa-Juss\`a and the anonymous reviewers for discussion and feedback. This work was supported by the KASP Scholarship Program and by the MINECO project HuMoUR TIN2017-90086-R.

\bibliography{emnlp-ijcnlp-2019}
\bibliographystyle{acl_natbib}

\end{document}